\documentclass[nonacm,sigplan,10pt]{acmart}
\renewcommand\footnotetextcopyrightpermission[1]{}
\AtBeginDocument{%
	}

\usepackage{amsmath}      
\usepackage[ruled,linesnumbered]{algorithm2e}
\usepackage{verbatim}     
\usepackage{textcomp}     
\usepackage{todonotes}
\usepackage{comment}
\usepackage{booktabs}
\usepackage{listings}
\usepackage{xcolor}
\usepackage{subcaption}
\usepackage{tikz}
\usepackage{float}
\usepackage{enumitem}
\usepackage{tabularx}
\usetikzlibrary{arrows.meta, positioning, decorations.pathreplacing, calc}
\newcommand{\gpumemutil}{\texttt{gpu\_\allowbreak mem\_\allowbreak util}}
\newcommand{\outputlen}{\texttt{output\_\allowbreak len}}

\begin{document}
	
	\title{vToken: Token-Level Virtualization for Reclaimable KV Caches}
	
	\author{Yuanhang Gao}
	\affiliation{%
		\institution{National University of Defense Technology}
		\city{Changsha}
		\country{China}}
	\email{gaoyh@nudt.edu.cn}

    \author{Xiangrui Yang}
	\affiliation{%
		\institution{National University of Defense Technology}
		\city{Changsha}
		\country{China}}
	\email{yangxiangrui11@nudt.edu.cn}

    \author{Yuanfeng Chen}
	\affiliation{%
		\institution{National University of Defense Technology}
		\city{Changsha}
		\country{China}}
	\email{chenyuanfeng@nudt.edu.cn}

    \author{Hongjia Chen}
	\affiliation{%
		\institution{National University of Defense Technology}
		\city{Changsha}
		\country{China}}
	\email{chenhongjia22@nudt.edu.cn}

    \author{Qianru Lv}
	\affiliation{%
		\institution{National University of Defense Technology}
		\city{Changsha}
		\country{China}}
	\email{lvqianru11@nudt.edu.cn}

    \author{Wenfei Wu}
	\affiliation{%
		\institution{Peking University}
		\city{Beijing}
		\country{China}}
	\email{wenfeiwu@pku.edu.cn}

    \author{Dongsheng Li}
    \authornote{Corresponding author.}
	\affiliation{%
		\institution{National University of Defense Technology}
		\city{Changsha}
		\country{China}}
	\email{dsli@nudt.edu.cn}
	
	\renewcommand{\shortauthors}{Gao et al.}
	
	\begin{abstract}
		Large language model serving faces a critical memory bottleneck: the KV cache grows with sequence length and batch size. PagedAttention uses fixed-size memory blocks to reduce allocator-level fragmentation, but recent KV eviction algorithms operate at a token granularity finer than block-level management. This mismatch causes intra-block fragmentation, leaving a large fraction of allocated KV memory unreclaimable.
		We present \textsc{vToken}, a lightweight token-level virtualization layer that decouples logical token liveness from physical block placement. \textsc{vToken} maintains a stable logical token view through token-table indirection and realizes physical reclamation by repacking live tokens asynchronously. The design preserves PagedAttention kernels and CUDA Graph compatibility.
		We implement \textsc{vToken} in vLLM and evaluate it with H2O, Random, and Scissorhands across models. Compared with a paired \textsc{Naive-Evict} baseline, \textsc{vToken} reduces retained KV blocks per request by 27.2\%--72.3\% and improves SLA-constrained throughput by up to 1.37$\times$. Under a constrained active-KV budget, it extends the maximum feasible concurrency by up to 2$\times$, while reducing the per-policy integration footprint from 500+ lines to under 50.
	\end{abstract}
	
	
	\settopmatter{printfolios=true,printacmref=false}
	\maketitle


	
	\section{Introduction}
	The rapid advancement of large language models (LLMs)~\cite{achiam2023gpt,liu2024deepseekv3,guo2025deepseekr1} has created an urgent demand for efficient inference serving systems. From conversational chatbots~\cite{chiang2023vicuna, chiang2024chatbot} to code generation~\cite{anthropic_claude_code_2026, openai_codex_2026, guo2024deepseekcoder}, LLMs are now deeply integrated into a wide spectrum of applications~\cite{openclaw_2026}. However, high-throughput, low-latency inference faces a critical bottleneck: the KV cache. During autoregressive generation, the KV cache stores the key and value tensors computed at each step for reuse in subsequent token generation, reducing attention complexity from \(O(N^2)\) to \(O(N)\). The size of the KV cache scales linearly with both context length and batch size. For long-context scenarios, KV cache can occupy tens of gigabytes of GPU memory, far exceeding the memory footprint of model weights. Consequently, KV cache management directly dictates the achievable concurrency and throughput of LLM serving systems~\cite{kwon2023efficient,sheng2023flexgen,aminabadi2022deepspeed}.
	
	PagedAttention~\cite{kwon2023efficient} and its implementation in vLLM have improved KV cache management by partitioning memory into fixed-size blocks, drawing inspiration from virtual memory paging~\cite{kilburn2009one} in operating systems. This block-based design reduces allocator-level fragmentation and enables cache sharing across requests. PagedAttention therefore serves as the \emph{block-level substrate} in our design: it provides efficient physical KV allocation and attention-kernel compatibility, while exposing only a block-granular interface for allocation and reclamation.
	
	Token-level KV eviction algorithms reduce memory pressure by exploiting the observation that not all historical tokens are equally important for future generation~\cite{xiao2023streamingllm}. A diverse family of strategies has emerged: H2O~\cite{zhang2023h2o} retains ``heavy hitter'' tokens with the highest cumulative attention scores; StreamingLLM~\cite{xiao2023streamingllm} preserves attention sink tokens plus a recent window; Scissorhands~\cite{liu2023scissorhands} and FastGen~\cite{ge2024fastgen} propose further criteria based on sparse attention patterns. However, although these techniques are effective individually, their naive composition with block-managed runtimes exposes a granularity mismatch: eviction policies decide which tokens should be retained or evicted, while the serving runtime maps, remaps, and reclaims KV memory at block granularity. This leaves a missing layer between token-level eviction decisions and block-level KV management.
	
	This mismatch is not merely a matter of wasted space, but a missing runtime boundary between token-level liveness and block-level reclamation. When a block contains both evicted and retained tokens, the block cannot be released, so its internal holes cannot be reused by other requests. Our preliminary evaluation shows that running token-level eviction on vLLM at a 16K context leaves most allocated blocks at no more than 50\% utilization across policies and workloads, with resulting intra-block waste exceeding 40\%. As a result, a substantial portion of GPU memory remains trapped in allocated but underutilized blocks, limiting effective serving capacity under memory pressure.
	Beyond fragmentation, this token/block granularity mismatch creates a barrier to algorithm integration. Integrating a new token-level eviction algorithm requires developers to understand block-manager internals and manually translate token-level retention semantics into block-level actions. This typically leads to two suboptimal outcomes: either waiting for an entire block to become empty, which sacrifices algorithm effectiveness, or modifying memory management behavior, which tightly couples policy logic with runtime code. This coupling slows algorithm iteration and makes it difficult to accommodate diverse future eviction strategies.
	
	A straightforward workaround is to reduce block size, for instance from 16 to 8 or 4 tokens. However, smaller blocks increase metadata overhead, address-mapping complexity, and the number of fine-grained non-contiguous KV transfers, which can reduce bandwidth efficiency in attention, copy, or offload paths~\cite{xie2025strata}. At the other extreme, pure token-level memory management---allocating space independently for each token---is impractical due to GPU memory allocation granularity and prohibitive metadata overhead. Thus, a new abstraction is needed that retains the efficiency of block-based management while supporting token-level semantics.
	
	Our core idea is to introduce \textsc{vToken}, a lightweight token-level memory virtualization layer between eviction policies and the PagedAttention substrate. This layer defines a clear semantic boundary: policies decide which tokens to evict, while the runtime maintains a token-level logical address space over block-managed KV memory. Upward, \textsc{vToken} exposes a token-level interface that lets policies operate without reasoning about physical blocks. Downward, it maintains logical-to-physical mappings, remaps attention slots, and reclaims physical blocks by asynchronously repacking live tokens when fragmentation becomes actionable. \textsc{vToken}'s primary contribution is the token-level virtualization abstraction; its physical reclamation backend realizes this boundary, using lazy compaction as the relocation mechanism in our implementation. \textsc{vToken} is a pressure-activated extension rather than a replacement for native full-KV serving. The native path remains preferable when KV memory is not the bottleneck; under memory pressure, \textsc{vToken} provides the runtime boundary that turns token-level KV eviction into reusable physical KV capacity.
	
	This paper makes the following contributions:
	\begin{itemize}[leftmargin=6pt, leftmargin=*]
		\item We identify the mismatch between token-level KV eviction and block-level KV cache management as a missing runtime abstraction layer and quantify its memory cost (\S\ref{subsec:mismatch}).
		\item We propose \textsc{vToken}, a token-level virtualization layer that decouples logical token liveness from physical block placement. \textsc{vToken} exposes a stable per-sequence token view upward and provides physical reclamation downward, without attention-kernel modifications.
		\item We implement \textsc{vToken} in vLLM and evaluate it across models and three eviction policies. \textsc{vToken} reduces retained KV blocks by 16\%--85\% and improves SLA-constrained throughput by up to 1.37$\times$ over \textsc{Naive-Evict}; under a constrained active-KV budget, it extends the maximum feasible concurrency by up to 2$\times$ and reduces per-policy integration from 500+ to under 50 lines of code.
	\end{itemize}
	
	\section{Background and Motivation}
	\label{sec:background}
	
	In this section, we provide the necessary background on KV cache management in LLM serving and token-level eviction algorithms. We then identify and quantify the granularity mismatch that motivates our work.
	
	\subsection{KV Cache and PagedAttention}
	\label{subsec:pagedattention}
	
	Modern autoregressive LLMs generate tokens sequentially. At each decoding step, the model computes attention over all previous tokens, requiring the Key and Value tensors of those tokens to be stored in GPU memory, i.e., the KV cache. For a model with \(L\) layers, \(H\) attention heads, and head dimension \(D\), the KV cache size per token is \(2 \times L \times H \times D \times \text{dtype\_size}\). For a 13B-parameter LLaMA model with 40 layers and 5120 hidden dimension, each token consumes approximately \(2 \times 40 \times 5120 \times 2\,\text{B} \approx 0.8\,\text{MB}\). With a 32K-token context, the KV cache alone exceeds 25 GB, dominating the memory footprint of inference.
	
	To manage this large cache efficiently, vLLM~\cite{kwon2023efficient} introduced \textsc{PagedAttention}, which organizes the KV cache into fixed-size \emph{blocks} (typically 16 tokens per block). PagedAttention maintains a block table that maps logical blocks to physical GPU memory blocks, analogous to virtual memory paging in operating systems. By allocating blocks on demand and allowing non-contiguous physical storage, it reduces external fragmentation; shared-prefix reuse further reduces memory consumption. As a result, PagedAttention has been widely adopted and serves as the foundation for many LLM inference engines \cite{tensorrt-llm, lightllm}.
	
	\subsection{KV Eviction Algorithms}
	\label{subsec:eviction}
	
	Recent works have proposed diverse token-level eviction strategies that exploit attention sparsity, each with different criteria for determining token importance: H2O~\cite{zhang2023h2o} (Heavy-Hitter Oracle) tracks the cumulative attention score of each token and retains only those with the highest scores---the ``heavy hitters,'' reporting up to 50\% cache reduction with limited perplexity increase. Sliding window policies retain only the most recent N tokens, discarding all earlier history. StreamingLLM~\cite{xiao2023streamingllm} preserves attention sink tokens (typically the first few tokens that accumulate high attention mass) plus a recent window. Random eviction probabilistically discards tokens to meet memory budgets. Scissorhands~\cite{liu2023scissorhands} and FastGen~\cite{ge2024fastgen} propose more sophisticated criteria based on attention patterns and token roles. Despite their algorithmic diversity, these approaches share a common requirement: they operate at the granularity of \emph{individual tokens}, making per-token retention decisions. When integrated into a serving system, they can reduce KV memory consumption and thereby support higher throughput or longer context windows.
	
	\subsection{The Granularity Mismatch Problem}
	\label{subsec:mismatch}
	
	Despite the benefits of both PagedAttention and token-level eviction, a fundamental mismatch exists between block-level memory management and token-level KV eviction. We first formalize this mismatch, then quantify its impact through preliminary experiments, and discuss its implications for algorithm integration.
	
	\subsubsection{Formal Definition}
	\label{subsubsec:frag_def}
	
	Let each block have capacity \(S\). For a physical block \(b\), let \(n_b\) denote the number of valid (non-evicted) tokens stored in that block, and define block utilization as \(u_b=n_b/S\). Given \(N\) allocated blocks, the intra-block waste ratio is
	\[
	F = 1 - \frac{\sum_{i=1}^{N} n_i}{N \cdot S} = 1 - \frac{1}{N}\sum_{i=1}^{N} u_i.
	\]
	
	When token-level eviction is applied, some tokens in a block become invalid, reducing \(n_b\) while the block itself remains allocated. This creates internal fragmentation: memory is reserved but underutilized.

	
	\begin{figure}[!t]
		\centering
		\setlength{\abovecaptionskip}{-1pt}
		\setlength{\belowcaptionskip}{-10pt}
		\includegraphics[width=\columnwidth]{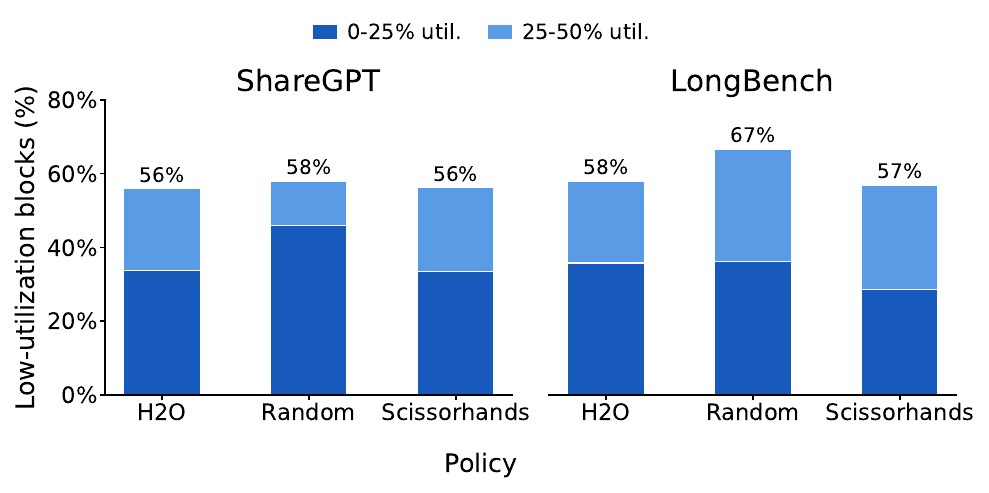}
		\caption{Low-utilization KV blocks under token-level eviction. Bars show allocated blocks with $\leq$50\% utilization, split into 0--25\% and 25--50\% bins; such partially live blocks remain unreclaimable in a block-granular runtime.}
		\label{fig:motivation}
	\end{figure}
	
	\subsubsection{Quantifying the Impact}
	\label{subsubsec:quantify}
	
	To understand the severity of this fragmentation, we conduct a preliminary experiment using vLLM integrated with a H2O-inspired token-level eviction algorithm (referred to as H2O hereafter). We run the Llama-3.1-8B model on a 16K-token subset of the ShareGPT~\cite{sharegpt2023} and LongBench~\cite{bai2024longbench} datasets with a batch size of 16. We measure the fraction of allocated blocks that remain low-utilization after token-level eviction.    
	
	\textit{Results.} Figure~\ref{fig:motivation} shows that token-level eviction leaves a large fraction of allocated blocks at low utilization. Across policies and workloads, most allocated blocks are at no more than 50\% utilization, and the resulting intra-block waste ratio \(F\) reaches 40--60\%. These blocks cannot be returned to the free pool because each still contains at least one retained token. Thus, token-level eviction reduces logical KV demand, but without a token-level virtualization layer, much of this reduction remains trapped as partially live physical blocks rather than reusable capacity.
	
	\subsubsection{Algorithm Integration Barrier}
	\label{subsubsec:barrier}
	
	Beyond memory waste, the granularity mismatch imposes a high cost on integrating new eviction algorithms into existing systems. Without a virtualization layer, a vLLM integration must compute token importance in the scheduler, select tokens to evict, translate those token-level decisions into block-level actions, and keep attention slot mappings consistent with any retained tokens. Because blocks cannot be partially freed, the implementer must either wait until an entire block becomes empty, delaying reclamation, or add custom logic to evacuate live tokens before releasing a block, which requires deep changes to the block allocator and careful synchronization.
	
	This process requires repeated engineering effort for each new algorithm and complicates production integration. In our experience, a direct integration of a single token-level eviction algorithm (e.g., H2O) into vLLM without a virtualization layer requires modifying over 500 lines of code across multiple core modules, with additional debugging effort to ensure correctness; each additional strategy requires comparable work.
	
	\section{vToken Design}
	\label{sec:design}
	
	\subsection{Design Goals and Virtualization Boundary}
	\label{subsec:goals}
	
	The granularity mismatch in \S\ref{sec:background} stems from a single missing piece: the runtime does not expose token-level liveness as a first-class concept. Two direct fixes are insufficient. A fully token-granular allocator would expose the right semantic unit, but it weakens the contiguous-access assumptions of PagedAttention kernels, multiplies per-entry metadata, and would require invasive CUDA changes across systems such as vLLM, TensorRT-LLM, and SGLang~\cite{zheng2024sglang}. Shrinking the block size redistributes the same fragmentation--bandwidth tradeoff without removing it. \textsc{vToken} therefore keeps the block-managed KV runtime intact and adds a thin \emph{virtualization boundary} that exposes token-granular semantics on top of it, leaving allocator and attention kernels untouched.
	
	The boundary defines a contract between two views of the same request. Above it, eviction policies operate on logical token identities: they emit \emph{which} tokens are no longer needed and never reason about physical blocks, attention slots, or reclamation timing. Below it, the runtime maintains a token-level logical address space, owns the logical-to-physical mapping, refreshes the slot mappings that attention kernels consume, and decides \emph{when} reclaiming fragmented blocks is profitable. Crucially, this contract decouples \emph{deciding} a token is dead from \emph{reclaiming} its KV memory, which lets reclamation be deferred and batched in the background.
	
	Maintaining this contract under live decoding raises three challenges that any realization of the boundary must solve. \emph{(C1)~Dual-view consistency.} A token may become dead independently of its neighbors, but a block can be released only when all useful KV entries inside it have been relocated or discarded; the runtime needs explicit metadata that simultaneously captures token liveness and block occupancy. \emph{(C2)~Safe reclamation during decoding.} Relocating KV entries changes the physical slots that attention reads, so slot mappings must be refreshed and asynchronous copies synchronized before any dependent attention kernel runs---without adding a global synchronization point to every decoding step. \emph{(C3)~Policy-neutral, amortized cost.} Different policies (H2O, StreamingLLM, Random) must reuse the same scheduler, block-manager, and worker-layout code, and indirection or reorganization cost must be paid only when fragmentation is actionable.
	
	\begin{figure}[!tbp]
		\centering
		\setlength{\belowcaptionskip}{-10pt}
		\includegraphics[scale=0.9]{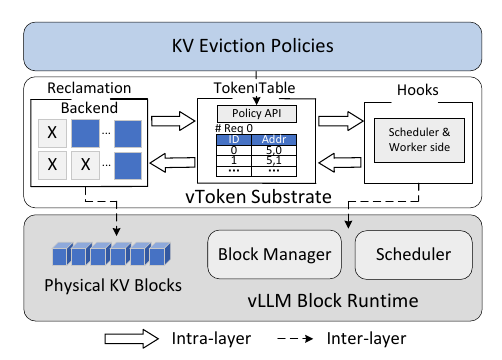}
		\caption{Overview of the \textbf{vToken} system}
		\label{fig:architecture}
	\end{figure}
	
	Figure~\ref{fig:architecture} shows how \textsc{vToken} answers each challenge: a per-sequence \emph{token table} provides the dual-view metadata for C1 (\S\ref{subsec:token-table}); a \emph{physical reclamation backend} batches reclamation in the background to satisfy C3 and to drive the asynchronous copies of C2 (\S\ref{subsec:compaction}); and \emph{reclamation-aware scheduler hooks} insert the slot-mapping refresh and CUDA-event dependency that close C2's safety requirement (\S\ref{subsec:scheduler}).
	
	\subsection{Token Table and Logical Address Space}
	\label{subsec:token-table}
	
	\begin{figure}[t]
		\centering
		\setlength{\belowcaptionskip}{-10pt}
		\includegraphics[scale=0.54]{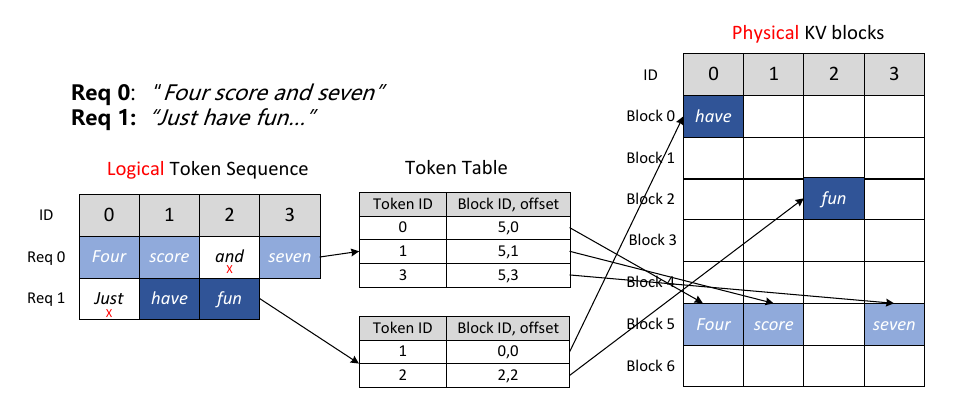}
		\caption{Token table as virtualization metadata for the logical token address space}
		\label{fig:token_table}
	\end{figure}
	
	The \textbf{token table} is the central metadata structure of \textsc{vToken} (Figure~\ref{fig:token_table}). It is maintained per request and records, for each logical token ID, the current physical location as a tuple (block ID, offset) along with a liveness bit. Marking a token reclaimable updates only this metadata; no KV memory is moved or freed until the reclamation backend acts. This indirection separates policy semantics (what to evict) from physical layout maintenance (how to repack live tokens).
	The table exposes two interfaces to upper-layer eviction policies, plus one internal interface used by the reclamation backend:
	
	\begin{itemize}[leftmargin=6pt, leftmargin=*]
		\item \texttt{evict\_token(req\_id, token\_id)}: Marks the token as logically inactive. The KV entry may remain physically present until reclamation, but it is excluded from subsequent attention slot mappings and treated as dead space by the reclamation planner.
		\item \texttt{sync\_new\_tokens(req\_id, block\_ids, total\_len)}: Registers newly generated tokens. The scheduler passes the request's current block list and updated sequence length; the table assigns sequential logical IDs and records their physical positions.
		\item \texttt{apply\_moves(req\_id, moves)}: Invoked by the reclamation backend after copies complete. Each move carries a token ID and its new (block, offset); the table atomically updates the affected entries so subsequent slot lookups see the post-relocation layout.
	\end{itemize}
	
	These interfaces are deliberately block-agnostic: a policy calls \texttt{evict\_token} without knowing which block holds the token or whether the block can be freed. This makes \textsc{vToken} a reusable substrate for diverse token-level policies. The metadata cost is modest---$O(L)$ per sequence of length $L$ (one mapping entry plus liveness state per token), independent of the number of layers, heads, or KV tensor elements; \S\ref{sec:implementation} reports the absolute footprint.	
	
	The canonical table lives on the CPU for simplicity and compatibility, while a GPU-resident lookup cache accelerates slot translation in the steady-state decoding path. After eviction or relocation, the cached translation array is refreshed before constructing the next attention slot mapping. When a sequence grows append-only, the cache is updated by appending the new entries; a full rebuild is triggered only on structural changes (eviction, relocation, or layout reconciliation).
	
	\subsection{Physical Reclamation: A Backend for the Virtualization Boundary}
	\label{subsec:compaction}
	
\begin{figure}[!t]
		\setlength{\belowcaptionskip}{-2pt}
		\noindent\includegraphics[width=\columnwidth]{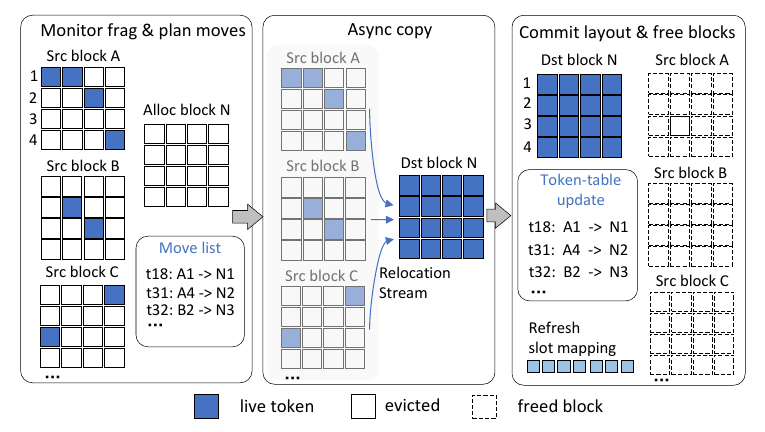}
		\caption{Physical reclamation workflow}
		\label{fig:compaction-steps}
	\end{figure}

	The reclamation backend realizes the virtualization boundary by translating logical token liveness into safe physical block reuse. In our implementation, lazy compaction is the concrete relocation mechanism: it repacks retained tokens from low-utilization blocks, updates the token table after copies complete, and returns emptied blocks to the allocator. Throughout relocation, the token table preserves a stable logical token view while physical KV locations change underneath. The backend consists of four stages---reclamation eligibility, headroom-aware admission, relocation planning, and asynchronous copy---as shown in Figure~\ref{fig:compaction-steps}.
	
	\subsubsection{Reclamation Eligibility}
	
	The backend uses per-block liveness metadata to decide when logical holes have become physically actionable, without exposing block state to the eviction policy. For each block \(b\), it maintains \(n_b\), the number of live tokens in that block (i.e., tokens not marked as evicted via \texttt{evict\_token}), and derives the intra-block waste ratio \(F\) defined in \S\ref{subsubsec:frag_def}. A high \(F\) indicates that many allocated blocks are underutilized.
	The metadata is updated incrementally on token eviction and block allocation/free events, so eligibility checks do not scan the full token table. To avoid excessive overhead, we amortize checks over multiple scheduling iterations and rate-limit reclamation attempts, so the planner is invoked only when fragmentation is likely to be actionable.
	
	\subsubsection{Headroom-Aware Admission}
	
	Out-of-place relocation needs destination blocks before fragmented source blocks can be released. Waiting until the free list is empty would therefore prevent reclamation from starting precisely when memory pressure is highest. \textsc{vToken} attempts reclamation only when the global intra-block waste ratio \(F\) exceeds a threshold \(\theta_F\), enough low-utilization blocks exist, and the free-block count is near a low watermark.
	The threshold \(\theta_F\) is tunable. A lower threshold attempts reclamation more frequently, potentially keeping fragmentation lower but incurring more overhead; a higher threshold reduces attempt frequency at the cost of higher average fragmentation. Through empirical exploration (\S\ref{subsec:sensitivity}), we set \(\theta_F = 0.25\) as a default, which balances memory efficiency and overhead in our experiments.
	The scheduler maintains bounded evacuation headroom inside the same KV block budget used for admission. This headroom is temporary working space for destination blocks, not an additional memory pool. When the free-block count approaches the low watermark, the scheduler prioritizes reclamation and applies admission backpressure instead of consuming the remaining free blocks with new requests. A reclamation plan is admitted only if the reserved headroom can provide the required destination blocks; otherwise, the plan is deferred until blocks become available.

	\subsubsection{Relocation Planning}
	
	Planning remains request-local. For each request, the backend selects eligible low-utilization source blocks, computes the live-token footprint, and constructs a relocation plan only when the projected block reduction is positive. The batch size \(B\) remains a configurable upper bound on the number of source blocks considered in one planning round. Let the selected source blocks contain a total of \(T_{\text{live}}\) live tokens.
	
	The backend then allocates destination blocks from the bounded evacuation headroom and the remaining free block pool. The number of destination blocks needed is \(\lceil T_{\text{live}} / S \rceil\). A plan is admitted only if these destinations are available and the source batch can release more blocks than it consumes as destinations; otherwise, the planner reduces the candidate set and emits no relocation plan for that round. This admission rule keeps reclamation profitable while preventing relocation from borrowing unbounded memory under pressure.
	The resulting move list iterates through the live tokens in the source blocks (preserving logical token order when possible to maintain cache locality) and assigns them sequentially to the destination blocks, filling each destination block completely before moving to the next. Each move records a token ID, its source location \((src\_block, src\_offset)\), and its destination location \((dst\_block, dst\_offset)\); applying the list updates the token table after copies complete.
	
	\begin{figure}[!t]
		\centering
		\setlength{\belowcaptionskip}{-2pt}
		\includegraphics[scale=0.5]{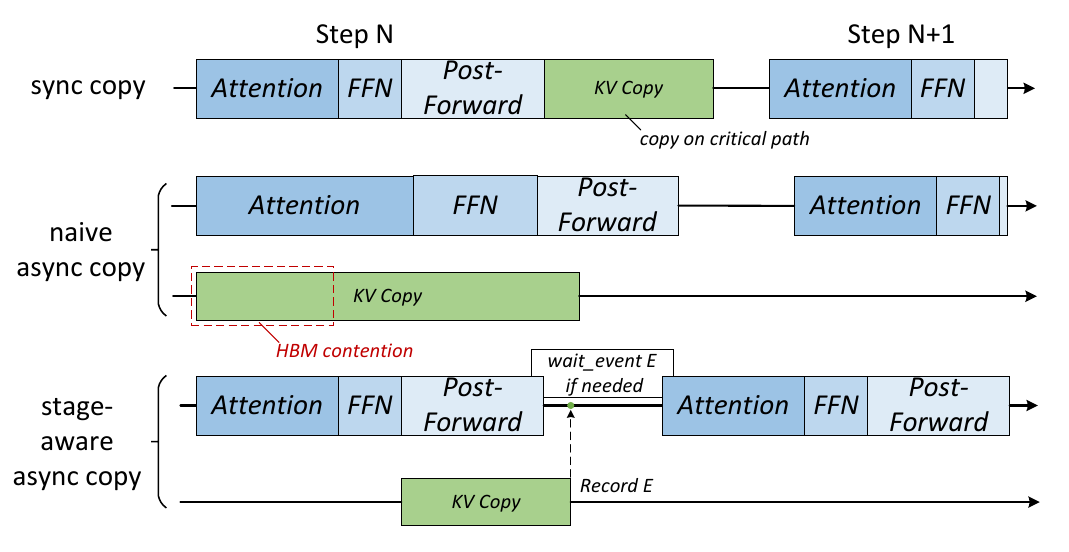}
		\caption{Stage-aware asynchronous copy. \textsc{vToken} avoids putting KV copy on the critical path and reduces HBM contention by launching copy after forward; a CUDA event guards the next attention step only when needed.}
		\label{fig:stage-aware}
	\end{figure}
	
	\subsubsection{Stage-Aware Asynchronous Copy}
	\label{sunsec:stage-aware}
	A decode step decomposes into three GPU phases with distinct resource profiles: (a) \emph{attention forward}, which reads the KV cache and is HBM-bandwidth bound; (b) \emph{FFN forward}, which is compute bound; and (c) \emph{post-forward work} (sampling, scheduling, next-step preprocessing), which is dominated by CPU-side logic with low GPU utilization. Na\"ively overlapping relocation copy with the entire step causes contention in phase~(a): both attention and KV-copy compete for HBM bandwidth, and the attention kernel slows down whenever a copy is in flight.
	
	As shown in Figure~\ref{fig:stage-aware}, \textsc{vToken} avoids this by launching KV copies \emph{after} the current step's forward returns, on a dedicated relocation stream. The current step's slot mappings have already been built from the old token table and consumed by the in-flight forward, so the post-forward token-table update cannot corrupt that step. The copies then proceed concurrently with sampling, scheduling, and next-step preprocessing---phases that do not read the KV cache and therefore do not contend with the relocation stream for HBM bandwidth. The next step's forward inserts a stream-level CUDA event dependency before model invocation; if the copy completes during post-forward work, the fence is a no-op. Because the dependency is expressed as a GPU-side \texttt{wait\_event} rather than a host synchronization, the CPU never stalls in the async path.
	
	Source blocks are returned to the free pool only after the new packed layout has been committed, and logical layout metadata is updated in the worker path after forward returns. This stage-aware schedule preserves the existing execution structure of vLLM while ensuring that subsequent attention steps observe a consistent post-relocation KV view without sharing HBM bandwidth with the attention kernel.
	
	\subsection{Integration with the Scheduler}
	\label{subsec:scheduler}
	
	Integrating \textsc{vToken} with vLLM's scheduler requires modifications in two key areas: slot mapping and scheduling hooks.
	
	\textbf{Slot mapping.} In vLLM, attention kernels access the KV cache via \emph{slots}, which are linearized indices computed as \(slot = block\_id \times block\_size + offset\). With the introduction of the token table, this direct computation is no longer valid because tokens may have been moved during relocation. Instead, we modify the slot mapping function to consult the token table:
	\[
	\begin{split}
		\text{slot} = \text{token\_table}[\text{token\_id}].block\_id \times S \\
		+ \text{token\_table}[\text{token\_id}].offset
	\end{split}
	\]
	This indirection makes attention slot construction follow the current physical location of each retained token after relocation. The lookup is performed once per token during input preparation rather than inside the attention kernel.
	
	\textbf{Scheduler hooks.} We integrate reclamation into vLLM's scheduling loop through a small set of scheduler- and worker-side hooks:
	\begin{itemize}[leftmargin=6pt, leftmargin=*]
		\item \textbf{Scheduler-side reservation hook}: Maintains the bounded evacuation headroom described in \S\ref{subsec:compaction} before admitting new KV allocations.
		\item \textbf{Worker-side reclamation hook}: During the worker execution path, the runtime evaluates reclamation opportunities, launches asynchronous KV-copy operations for selected requests, and rewrites the corresponding logical layout metadata.
		\item \textbf{Pre-attention synchronization hook}: Before attention consumes KV entries that may have been relocated, the worker inserts a stream-level dependency on the relocation event, ensuring that the moved KV data are visible to subsequent kernels.
	\end{itemize}
	
	\textbf{Compatibility with CUDA Graph.} \textsc{vToken} does not disable CUDA Graphs or recapture graphs after relocation. The KV cache tensors and graph-captured input buffers remain stable; only the contents of the mutable slot-mapping buffer are updated before replay according to the token table. Relocation runs outside the captured graph on a separate CUDA stream, and the worker inserts a CUDA-event dependency before any replay that may read relocated KV entries. Thus, \textsc{vToken} preserves CUDA Graph execution through dynamic slot-mapping adjustment rather than graph bypass or graph recapture.

	\subsection{Correctness Invariants}
	\label{subsec:consistency}
	
	The components in \S\ref{subsec:token-table}--\S\ref{subsec:scheduler} are coordinated by four invariants that together define what ``correctness under background relocation'' means in \textsc{vToken}. The mechanisms enforcing each invariant have been introduced above; here we state them explicitly so that the validation experiments in \S\ref{subsec:logicalkv_consistency} can be read as direct tests.
	
	\begin{itemize}[leftmargin=6pt, leftmargin=*]
		\item \textbf{I1 (Token conservation).} For each request, the set of active logical token IDs and their associated K/V tensor contents are preserved across every relocation event. Relocation may change the physical (block, offset) of a token but never its logical identity or its KV payload.
		\item \textbf{I2 (Unique mapping).} Every active logical token maps to exactly one (block, offset), and every occupied physical slot is referenced by at most one logical token. The token table enforces this on \texttt{apply\_moves} by atomically clearing the source slot and writing the destination slot.
		\item \textbf{I3 (Pre-attention visibility).} A relocated KV entry is read by attention only after the corresponding copy has completed. For each committed relocation, the worker records a CUDA event on the copy stream and inserts a stream-level wait before any subsequent attention launch whose slot mapping may reference the relocated entries. This dependency replaces a global GPU synchronization.
		\item \textbf{I4 (Layout-aware planning).} A relocation plan is committed only from a consistent block-list and liveness snapshot of the affected requests, never overlaps an in-flight plan on the same request, and is rejected when the projected post-relocation block count does not strictly decrease. Shared-prefix blocks are excluded from plans and handled conservatively as discussed in \S\ref{sec:discussion}.
	\end{itemize}
	
	Invariants I1 and I2 are checked online for every relocation event during evaluation (\S\ref{subsec:logicalkv_consistency}). I3 is structural: it follows from the CUDA-event topology established in \S\ref{subsec:compaction}--\S\ref{subsec:scheduler} and is asserted in unit tests rather than measured. I4 is enforced by the planner's profitability and isolation gates described in \S\ref{subsec:compaction}.
	
	\section{Implementation}
	\label{sec:implementation}
	
	Our \textsc{vToken} prototype is implemented on top of vLLM (v0.18.0), using PyTorch v2.10.0. It adds three components: a per-sequence \texttt{TokenTable} for logical-to-physical mappings, a \texttt{ReclamationManager} for fragmentation monitoring and relocation planning, and a \texttt{CUDACopyEngine} for asynchronous KV movement on a separate CUDA stream. Token-table space is separate from these runtime costs. The table is maintained per sequence rather than per layer; even with a conservative 16-byte entry, a 16K-token sequence requires only 256\,KB of metadata, below 0.1\% of the roughly 2\,GB FP16 KV cache footprint for the evaluated 7B/8B models.
	
	The vLLM-based prototype requires scheduler and worker hooks for headroom reservation, slot mapping, relocation launch, and pre-attention synchronization. The worker maintains cached slot-translation arrays and refreshes them only after structural layout changes. KV relocation is launched on a non-blocking CUDA stream, while the main stream waits on CUDA events before any attention launch that may consume relocated entries. The prototype targets the single-node, single-GPU decoding fast path, isolating the interaction between token-level eviction and block-managed KV allocation without conflating it with distributed scheduling or cross-device KV movement. Per-block liveness metadata let the planner find underutilized blocks without scanning the full table; a greedy packer with conservative gating triggers relocation only when the projected gain exceeds a threshold.
	
	With this runtime layer in place, policy integration becomes token-centric rather than block-centric. A policy only needs to report newly appended tokens and selected victims; for example, H2O calls \texttt{sync\_new\_tokens()} for generated tokens and \texttt{evict\_token()} for tokens whose scores fall below its budget. \textsc{vToken} then updates the token table, refreshes slot mappings, and schedules physical reclamation without requiring the policy to manipulate vLLM blocks directly.
	
	\textbf{Integration complexity.}
	Figure~\ref{fig:vtoken-policy} summarizes the resulting adapter path and code footprint: policies define only \textsc{SelectVictims}, while token-table updates, slot remapping, and reclamation are shared runtime functions.
	
	\begin{figure}[!h]
		\centering
		\footnotesize
		\setlength{\tabcolsep}{3pt}
		\renewcommand{\arraystretch}{1.05}
		\begin{tabular}{@{}p{0.19\columnwidth} p{0.48\columnwidth} p{0.28\columnwidth}@{}}
			\toprule
			\multicolumn{3}{@{}l}{\textbf{\textsc{vToken} policy-adapter procedure}} \\
			\midrule
			1 & \textsc{SyncNewTokens}$(r, B_{\mathrm{new}}, L)$ & shared adapter \\
			2 & $V \leftarrow P.\textsc{SelectVictims}(r)$ & policy hook \\
			3 & \textbf{for each} $t \in V$: \textsc{EvictToken}$(r,t)$ & shared adapter \\
			4 & $S \leftarrow \textsc{BuildSlotMapping}(r)$ & shared runtime \\
			5 & \textsc{ReclaimAsync}$(r)$ & shared runtime \\
			6 & \textsc{Decode}$(r,S)$ & worker \\
			\midrule
			\multicolumn{3}{@{}p{0.92\columnwidth}@{}}{\textit{Policy specialization.} H2O selects lowest-score tokens; Scissorhands retains tokens with persistent attention patterns and evicts low-persistence tokens; Random samples from evictable tokens.} \\
			\midrule
			\multicolumn{3}{@{}l}{\textit{Integration footprint.}} \\
			Block-native & scheduler/worker & \mbox{4--6 files, 500+ LOC} \\
			\textsc{vToken} & policy adapter & \mbox{1--2 files, <50 LOC} \\
			\bottomrule
		\end{tabular}
		\caption{Policy-adapter procedure and integration footprint in \textsc{vToken}.}
		\label{fig:vtoken-policy}
		\vspace{-2pt}
	\end{figure}
	\begin{figure*}[t]
		\centering
		\setlength{\belowcaptionskip}{-2pt}
		\includegraphics[scale=0.38]{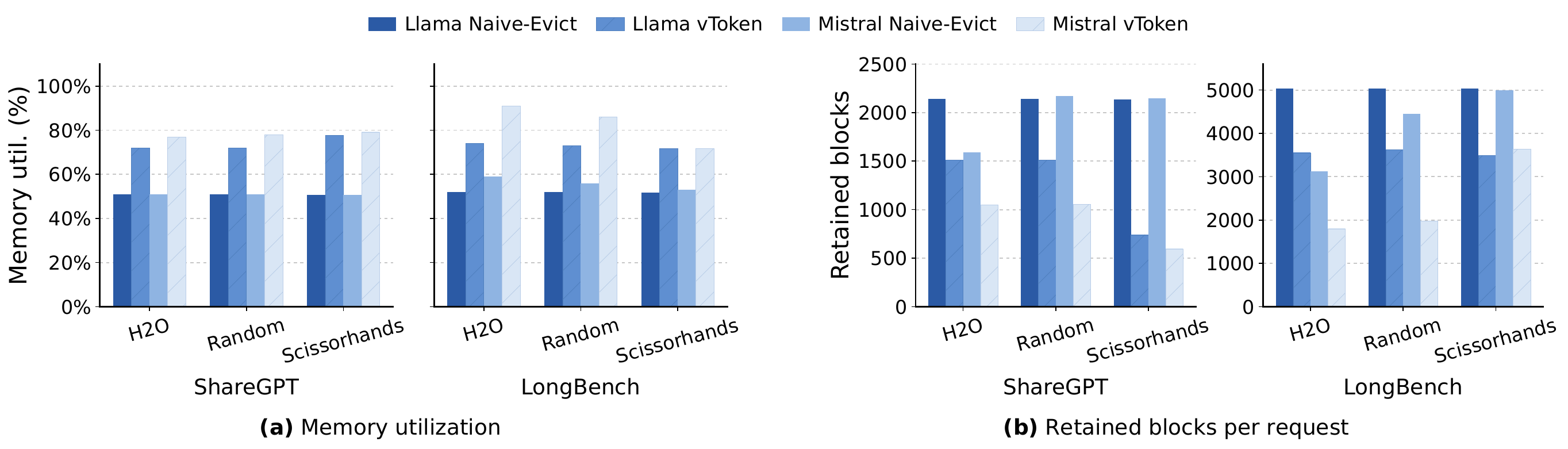}
		\caption{Memory efficiency under token-level eviction. \textsc{vToken} improves effective KV cache usage by reclaiming partially live blocks that remain allocated in \textsc{Naive-Evict}.}
		\label{fig:memory-efficiency}
	\end{figure*}
	\begin{figure*}[t]
		\centering
		\setlength{\belowcaptionskip}{-2pt}
		\includegraphics[scale=0.43]{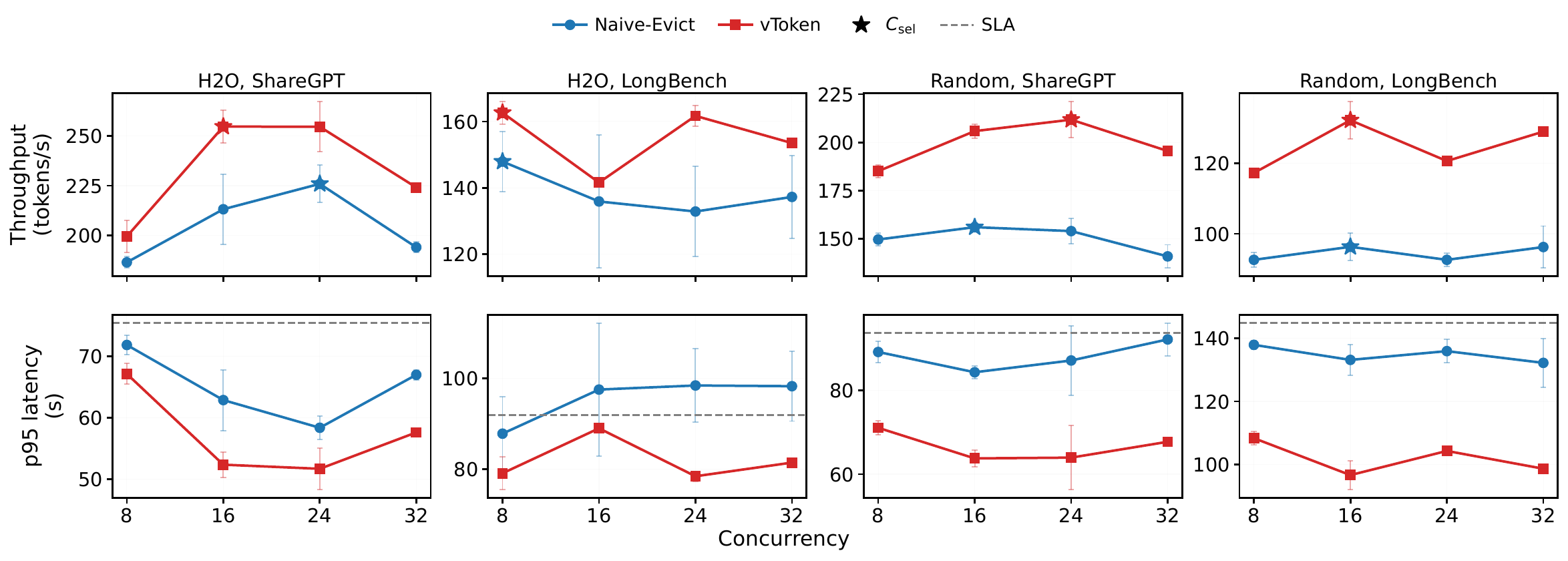}
		\caption{SLA-constrained throughput frontier on Mistral-7B under token-level eviction. The SLA threshold is 1.05$\times$ the p95 latency of \textsc{Naive-Evict} at $C_{\mathrm{ref}}=8$, and the star marks $C_{\mathrm{sel}}$, the maximum-throughput feasible point.}
		\label{fig:e2_frontier_mistral}
	\end{figure*}
	
	\section{Evaluation}
	\label{sec:evaluation}
	Our evaluation answers six questions: whether token-level eviction leaves physical capacity trapped in partially live blocks (\S\ref{subsec:memory_efficiency}), whether \textsc{vToken} improves the paired eviction frontier under identical token-level decisions (\S\ref{subsec:throughput}), whether it extends the active-KV capacity frontier under memory pressure (\S\ref{subsec:capacity}), where overhead and overlap costs arise (\S\ref{subsec:overhead}), how sensitive the system is to runtime parameters (\S\ref{subsec:sensitivity}), and whether it remains compatible with production serving features such as prefix caching while preserving relocation correctness (\S\ref{subsec:prefix_cache}--\S\ref{subsec:logicalkv_consistency}).	
	
	\subsection{Experimental Setup}
	\label{subsec:setup}
	
	\textbf{Platform, workloads, and baselines.} We evaluate \textsc{vToken} on a single NVIDIA H100 GPU with 80\,GB of memory. The main evaluation uses Mistral-7B and Llama-3.1-8B on ShareGPT and LongBench workloads; \S\ref{subsec:capacity} additionally reports a Qwen2.5-14B capacity-frontier check. We consider three token-level KV eviction policies: H2O, Scissorhands, and Random, and distinguish three system variants. \textsc{Native vLLM} is the unmodified full-retention serving baseline without \textsc{vToken} hooks or token-level eviction. \textsc{Naive-Evict} applies the same token-level eviction decisions on the block-based runtime but disables \textsc{vToken}'s physical reclamation backend, so partially live blocks remain allocated. \textsc{vToken} enables token-table indirection and physical reclamation on top of the same eviction policies. All reported experiments run with CUDA Graph enabled.
	
	\textbf{Methodology.} For paired comparisons between \textsc{Naive-Evict} and \textsc{vToken}, both variants use the same prompts, decoding settings, model, policy, and memory budget; the only difference is whether token-table indirection and physical reclamation are enabled. We set the decoding temperature to 0 and use \texttt{gpu\_mem\_util=0.90} except in the capacity-frontier experiment, which uses controlled KV budgets as explained in \S\ref{subsec:capacity}. Throughput experiments use a closed-loop concurrency scan with the SLA threshold defined from the \textsc{Naive-Evict} baseline at a reference concurrency. The vLLM version used in our prototype requires block sizes of at least 16 tokens, so the block-size sweep in \S\ref{subsec:sensitivity} is restricted accordingly.  
	
	\subsection{Memory Efficiency}
	\label{subsec:memory_efficiency}
	We compare \textsc{Naive-Evict} and \textsc{vToken} across all model--dataset--policy combinations. This paired comparison isolates physical reclamation: both variants apply the same token-level eviction decisions. We report two complementary metrics: memory utilization measures how effectively allocated KV blocks store retained tokens, while retained blocks per request measure the physical KV footprint that determines admission headroom under memory pressure.
	
	Figure~\ref{fig:memory-efficiency}a shows that \textsc{vToken} improves memory utilization by translating token-level liveness information into physical block reclamation. Compared with \textsc{Naive-Evict}, \textsc{vToken} increases average memory utilization by 21.88\% on Llama-3.1-8B and by 21.67\% on Mistral-7B. The gain appears consistently across workloads and policies, indicating that the effect is not tied to a single model or prompt-length distribution.
	As shown in Figure~\ref{fig:memory-efficiency}b, \textsc{vToken} also reduces the number of retained blocks by 27.2\%--72.3\% relative to \textsc{Naive-Evict}. This block-level reduction is the key systems effect: in \textsc{Naive-Evict}, a block remains allocated as long as it contains any retained token, so token-level holes do not become reusable capacity. \textsc{vToken} compacts the remaining live tokens into fewer physical blocks and returns emptied blocks to the allocator.
	These results show that the main loss in \textsc{Naive-Evict} is not the eviction policy itself, but the inability of a block-granular runtime to turn token-level liveness into reusable physical capacity. Retained blocks are the more operational metric: they directly determine how many active requests can remain resident before the KV block pool becomes the admission bottleneck. Because decoding is typically memory-bound, this reclaimed KV memory can support more concurrent requests. We evaluate this system-level effect next.    
	
	\vspace{-6pt}
	\subsection{Sustainable Throughput under Token-Level Eviction}
	\label{subsec:throughput}
	
	We evaluate whether the block-level capacity reclaimed in \S\ref{subsec:memory_efficiency} translates into SLA-constrained throughput improvement. The SLA threshold is $1.05\times$ the \textsc{Naive-Evict} p95 latency at $C_{\mathrm{ref}}=8$: anchoring to the paired baseline gives both variants the same latency budget, and the 5\% slack admits minor tail-latency variance without absorbing regressions. Figure~\ref{fig:e2_frontier_mistral} plots Mistral-7B; Llama-3.1-8B results are summarized below. Across all plotted workload--policy combinations, \textsc{vToken} consistently shifts the feasible frontier upward at \(C_{\mathrm{sel}}\).
	
	On Mistral-7B, \textsc{vToken} improves selected feasible throughput by 9.9\%--37.3\% and reduces p95 latency by 9.9\%--27.5\%. The largest gains occur under Random eviction, where live tokens are scattered across blocks and \textsc{Naive-Evict} retains many partially live blocks; H2O yields smaller but consistently positive gains because its retained tokens are more structured. We observe the same trend on Llama-3.1-8B: across six workload--policy combinations, \textsc{vToken} improves selected feasible throughput by 18.9\% on average and reduces p95 latency by 14.7\%. The gain is strongest under Random eviction, reaching 37.0\% on ShareGPT and 35.9\% on LongBench, while H2O still shows gains of 12.5\% and 5.7\%. Scissorhands shows stronger gains: across Mistral-7B and Llama-3.1-8B, \textsc{vToken} improves selected feasible throughput by 33.3\%--103.7\% and reduces p95 latency by 21.8\%--33.0\%. This is expected because Scissorhands retains tokens according to persistent attention patterns, leaving live tokens more dispersed across physical blocks.
	
	Figure~\ref{fig:e2_frontier_mistral} marks \(C_{\mathrm{sel}}\), the feasible point with the highest throughput under the SLA constraint. The selected point need not be the largest tested concurrency: the gains reported above are realized at \(C_{\mathrm{sel}}\), confirming that \textsc{vToken} improves the operating frontier rather than merely pushing to larger batches. Thus, the SLA-frontier gain comes from reclaiming capacity that already exists logically after eviction but remains physically trapped in partially live blocks without \textsc{vToken}.
	
	\begin{figure}[t]
		\centering
		\setlength{\belowcaptionskip}{-12pt}
		\includegraphics[width=0.9\columnwidth]{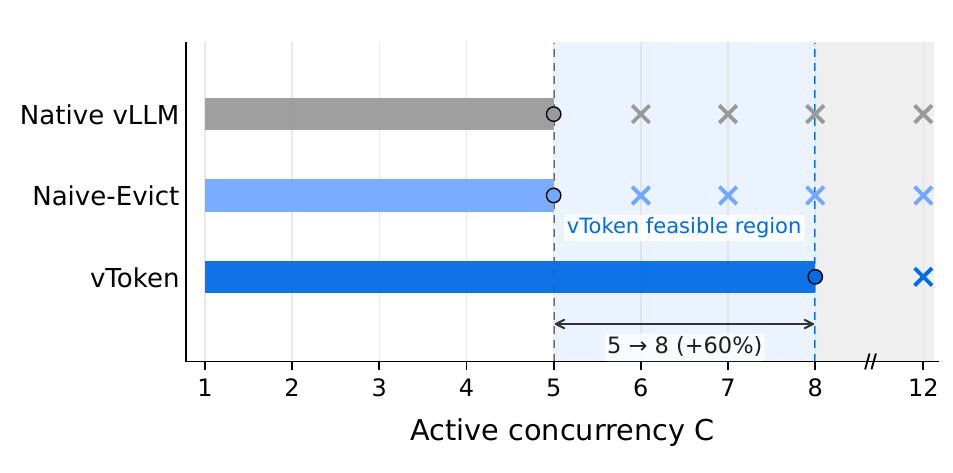}
		\caption{Active-KV capacity frontier at \texttt{gpu\_mem\_util=0.35}. \textsc{vToken} extends the verified feasible region from \(C=5\) to \(C=8\); \textsf{\texttimes} marks infeasible points.}
		\label{fig:capacity_frontier}
	\end{figure}
	
	\begin{figure}[t]
		\centering
		\begin{minipage}[t]{0.49\columnwidth}
			\centering
			\includegraphics[width=\linewidth]{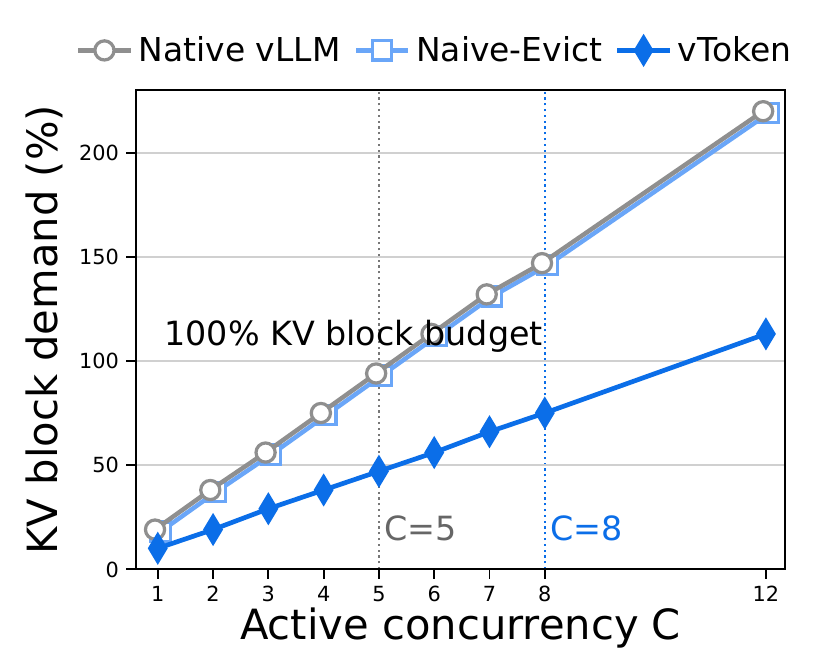}
		\end{minipage}
		\hfill
		\begin{minipage}[t]{0.49\columnwidth}
			\centering
			\includegraphics[width=\linewidth]{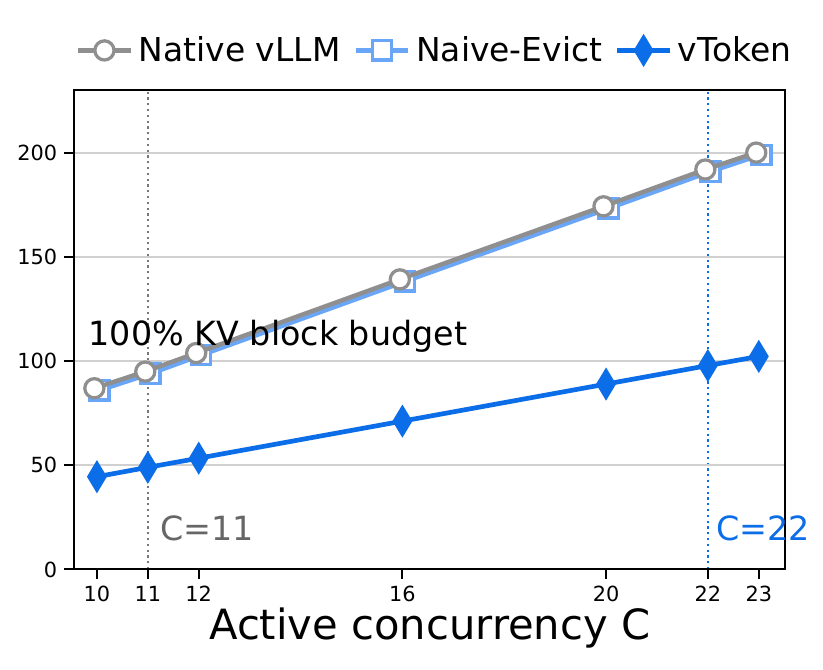}
		\end{minipage}
		\caption{KV-block demand normalized by usable budget. At \gpumemutil=0.35 (left) and 0.50 (right), \textsc{vToken} extends the verified boundary from $C=5$ to $C=8$ and from $C=11$ to $C=22$.}
		\label{fig:kv-demand}
	\end{figure}
	
	\subsection{Capacity Frontier under Memory Pressure}
	\label{subsec:capacity}
	
	\begin{figure*}[t]
		\centering

		\subfloat[Overhead composition.]{%
			\raisebox{0.6em}{%
				\includegraphics[width=0.31\textwidth]{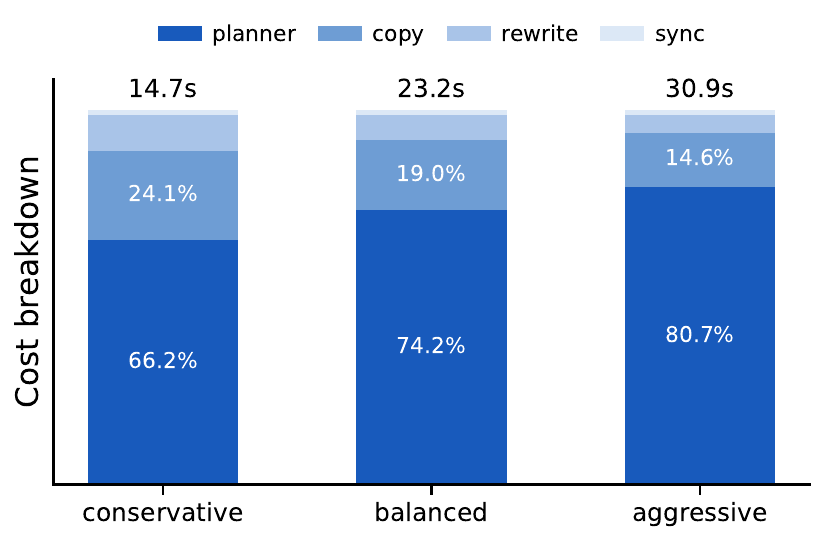}}%
			\label{fig:e4-overhead-composition}}
		\hfill
		\subfloat[Per-copy KV relocation cost.]{%
			\includegraphics[width=0.31\textwidth]{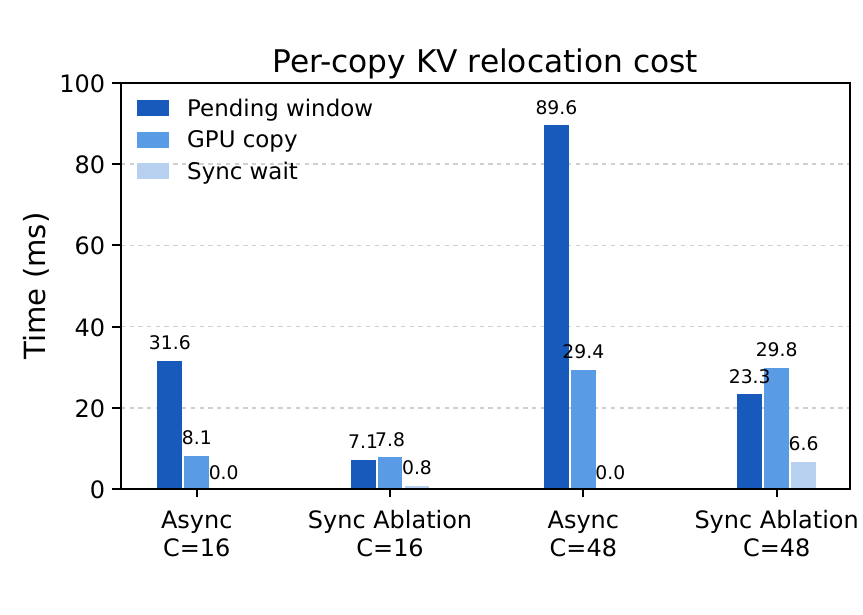}%
			\label{fig:e4a-cost}}
		\hfill
		\subfloat[Throughput and decode p95.]{%
			\includegraphics[width=0.31\textwidth]{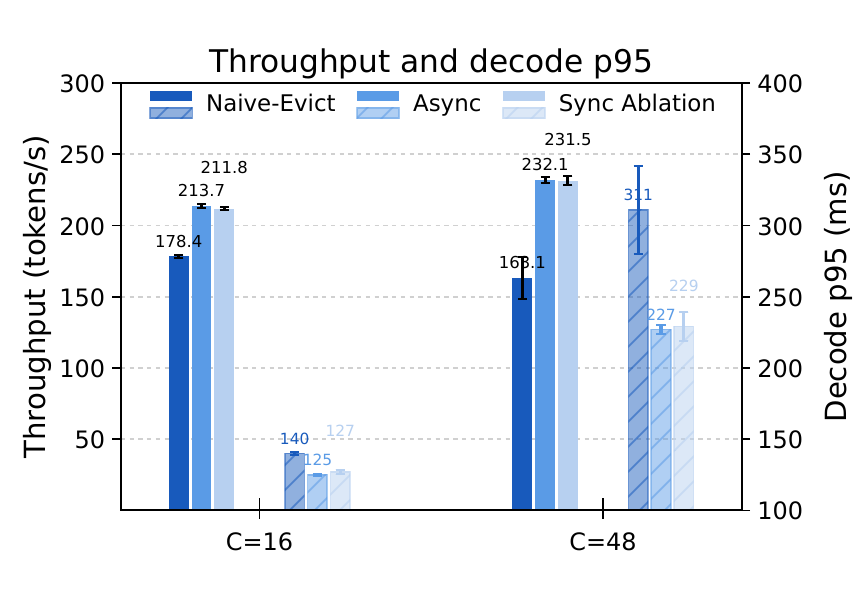}%
			\label{fig:e4a-throughput}}
		
		\caption{\textbf{Physical reclamation overhead and overlap.}
			(a) Planner-side work dominates CPU-observed overhead.
			(b) Async copies avoid explicit synchronization waits.
			(c) Async reclamation improves throughput and decode p95 over \textsc{Naive-Evict} while staying close to force-sync without explicit waits.}
		\label{fig:e4-overhead-overlap}
	\end{figure*}
	
	This experiment asks whether physical reclamation increases the number of active requests that can be simultaneously resident under the same KV block budget. It isolates the active-KV capacity frontier rather than vLLM's admission queue: a point is infeasible when the target active requests cannot fit in the active KV block pool. We compare \textsc{Native vLLM}, \textsc{Naive-Evict}, and \textsc{vToken} under the same allocator budget; \textsc{vToken}'s evacuation headroom is charged against the same usable block pool rather than allocated as extra memory. For feasible points, throughput is averaged over three runs.
	
	We use Llama-3.1-8B with LongBench, H2O, and \outputlen=12288. The H100 exposes 5,427 usable KV blocks at \gpumemutil=0.35 and 11,519 at 0.50. In this setup, a fully retained request needs 1,020 blocks, whereas the ideal packed footprint after H2O retention is 512 blocks. More generally, for active context length \(L\), block size \(B\), and \(R(L)\) retained tokens after eviction, full retention scales as \(C\lceil L/B\rceil\) blocks, while an ideally packed token-evicted cache scales as \(C\lceil R(L)/B\rceil\). \textsc{Naive-Evict} tends toward the former because partially live blocks remain allocated; \textsc{vToken} moves toward the latter by turning retained tokens into reclaimable physical blocks.
	
	Figure~\ref{fig:capacity_frontier} reports the capacity frontier, while Figure~\ref{fig:kv-demand} explains the underlying block-demand mechanism. At \gpumemutil=0.35, \textsc{Native vLLM} and \textsc{Naive-Evict} are feasible only through \(C=5\). Six full-retention requests would require 6,120 blocks and exceed the 5,427-block pool, and \textsc{Naive-Evict} remains close to this demand because live tokens stay scattered across partially live blocks. \textsc{vToken} closes the gap toward the 512-block packed footprint, remains feasible through \(C=8\), and extends the maximum feasible concurrency by \(\mathbf{60\%}\). Its boundary throughput at \(C=8\) is 180.3\,tokens/s, close to its own peak of 203.2\,tokens/s at \(C=5\), showing graceful degradation rather than collapse.
	
	The same mechanism persists under the larger controlled budget. At \gpumemutil=0.50, \textsc{Native vLLM} and \textsc{Naive-Evict} reach their verified frontier at \(C=11\), whereas \textsc{vToken} remains below the usable KV-block budget through \(C=22\), doubling the verified feasible concurrency. Figure~\ref{fig:kv-demand} shows that the shift comes from lower per-request physical block demand under the same allocator budget, not from extra headroom or a different admission policy.
	
	\textbf{Larger-model capacity check.} We repeat the capacity-frontier sweep on Qwen2.5-14B with \outputlen=8192 and \gpumemutil=0.50, leaving 3,261 usable KV blocks. \textsc{Native vLLM} and \textsc{Naive-Evict} are feasible only through \(C=3\) and first fail at \(C=4\), whereas \textsc{vToken} remains feasible through \(C=6\) and fails at \(C=8\). This \(2\times\) frontier extension indicates that the capacity benefit is not specific to Llama-3.1-8B. This check isolates capacity-frontier extension: native full-KV serving remains preferable when feasible, while \textsc{vToken} extends the feasible region once full retention exhausts the KV block budget. The sensitivity of this boundary to the eviction ratio is examined in \S\ref{subsec:sensitivity}.

	\subsection{Overhead and Overlap}
	\label{subsec:overhead}
	
	This subsection separates CPU-observed runtime overhead from GPU data-movement behavior and checks whether the physical reclamation backend overlaps with decoding. We run this mechanism experiment on Llama-3.1-8B with LongBench and H2O. We first verify that the steady-state indirection path is nearly free: an indirection-only ablation that installs the \textsc{vToken} hooks but disables eviction and reclamation, with slot translation refreshed only on structural changes, changes throughput and p95 by less than 1.0\% relative to \textsc{Native vLLM} at both C=16 and C=48. The overhead reported below therefore comes from the pressure-activated planning and relocation backend, not from per-attention table lookup. Figure~\ref{fig:e4-overhead-overlap}(a) reports planner work, host-side copy launch/accounting, and token/block-table rewrites across 100 profiling runs. Planning consistently dominates the CPU-observed overhead, indicating that the main measured cost is planner-side opportunity checking rather than KV relocation accounting.
	
	Figure~\ref{fig:e4-overhead-overlap}(b) validates the asynchronous path with a force-synchronization ablation. In the normal path, relocation copies incur no explicit synchronization wait: the copy remains pending for multiple decode steps, while GPU copy time is much shorter than the pending window. In contrast, the force-sync ablation exposes CPU blocking per relocation event, confirming that async overlap avoids an explicit synchronization stall.
	
	The overlap is still not free, since copy kernels share GPU resources with decode. Figure~\ref{fig:e4-overhead-overlap}(c) shows that async reclamation improves both throughput and decode p95 over \textsc{Naive-Evict} at \(C=16\) and \(C=48\), while staying close to force-sync without exposing synchronization stalls. Since reclamation is less frequent at the capacity-frontier operating points, these measurements should be read as a stress test of the overlap mechanism rather than the steady-state cost at the frontier. Thus, \textsc{vToken} removes explicit stalls and overlaps copy with decode, but copy/decode contention can appear under heavier operating pressure.
	
	\subsection{Sensitivity Analysis}
	\label{subsec:sensitivity}
	
\begin{figure}[t]
		\setlength{\belowcaptionskip}{-2pt}
		\noindent\includegraphics[scale=0.53]{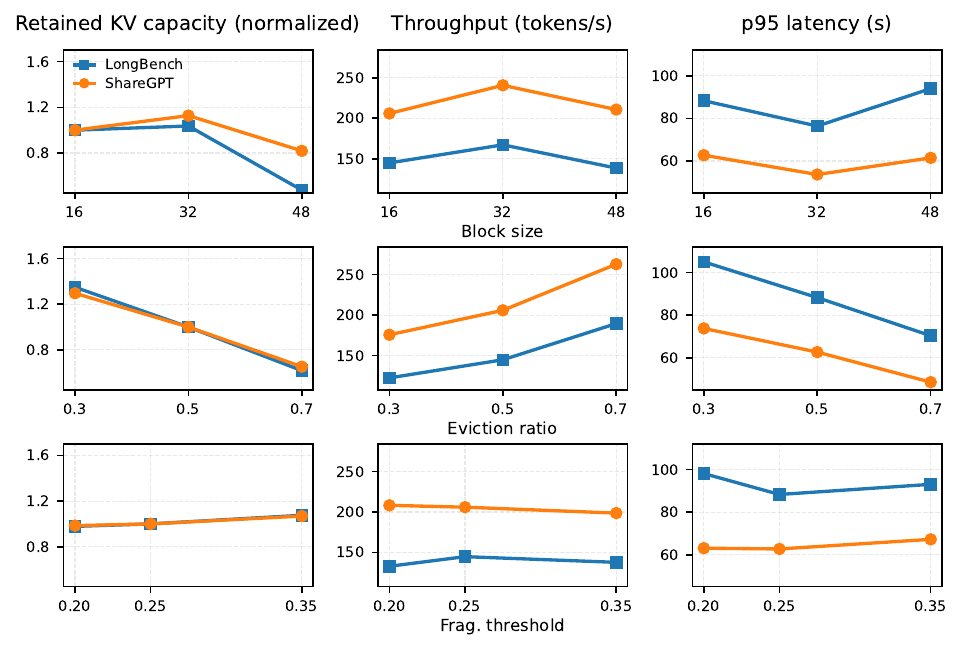}
		\caption{Sensitivity analysis of block size, eviction ratio, and fragmentation threshold under H2O.}
		\label{fig:e3-sensitivity}
	\end{figure}

	We use one-factor-at-a-time experiments to evaluate how \textsc{vToken} responds to three runtime parameters: block size, eviction ratio, and fragmentation threshold. Figure~\ref{fig:e3-sensitivity} reports normalized retained KV capacity, throughput, and p95 latency under fixed concurrency for H2O on LongBench and ShareGPT. The retained KV capacity is computed as retained blocks multiplied by block size, and is normalized to the default configuration for each model--dataset pair. We show H2O as the representative policy because StreamingLLM and Random exhibit the same qualitative trends across these sweeps.
	
	\textbf{Block size.} We start from block size 16 because it is the minimum supported by vLLM's PagedAttention allocator in our evaluated configuration. Changing the block size affects both allocation granularity and the physical capacity represented by each block. Normalized retained KV capacity therefore provides a comparable view across block sizes. Both workloads achieve the best throughput and lowest p95 latency at block size 32, while block size 48 reduces retained capacity but degrades performance. This shows that simply enlarging blocks creates a capacity/performance tradeoff and is not a substitute for token-level reclamation.
	
	\textbf{Eviction ratio.} Increasing the eviction ratio exposes more reclaimable KV capacity. As the ratio increases from 0.3 to 0.7, normalized retained KV capacity decreases substantially on both workloads, while throughput improves and p95 latency drops consistently. This shows that \textsc{vToken} can convert more aggressive token eviction into physical KV-cache reclamation. This sweep stresses the reclamation mechanism; selecting an accuracy-preserving eviction budget remains the responsibility of the underlying eviction policy.
	
	\textbf{Fragmentation threshold.} The fragmentation threshold is less dominant than the eviction ratio. Raising the threshold delays reclamation slightly, which modestly increases normalized retained KV capacity and mildly affects p95 latency, while throughput remains comparatively stable across the evaluated range. This indicates that \textsc{vToken} does not require fine-grained trigger tuning: the main effect comes from exposing token-level holes, while the reclamation trigger controls when physical reclamation is worthwhile.
	
	\subsection{Prefix-Cache Compatibility}
	\label{subsec:prefix_cache}
	\begin{table}[tb]
		\centering
		\caption{Prefix-cache compatibility.}
		\vspace{-0.2em}
		\label{tab:prefix-cache}
		\scriptsize
		\setlength{\tabcolsep}{2.2pt}
		\renewcommand{\arraystretch}{0.96}
		\begin{tabular}{lrrrrr}
			\toprule
			Degree & Shared skipped & Suffix reclaimed & Block red. & Throughput $\Delta$ & p95 $\Delta$ \\
			\midrule
			1 (no sharing) & n/a & 535K & 28.6\% & -2.9\% & +2.9\% \\
			2 & 100\% & 65.6K & 33.0\% & +76.1\% & -40.2\% \\
			4 & 100\% & 198K & 32.3\% & +97.2\% & -40.9\% \\
			8 & 100\% & 264K & 42.2\% & +146.5\% & -47.8\% \\
			\bottomrule
		\end{tabular}
		\vspace{-2pt}
	\end{table}
	
	Prefix caching is a common production serving feature. \textsc{vToken}'s current implementation conservatively excludes shared-prefix blocks from relocation while keeping private suffix blocks eligible for reclamation. We validate this boundary on Llama-3.1-8B using synthetic H2O workloads with an 8K-token shared prefix, an 8K-token private suffix, and sharing degrees from 1 to 8.
	
	Table~\ref{tab:prefix-cache} shows that for sharing degrees 2, 4, and 8, all shared-prefix candidates are skipped and all consistency checks pass. Relative to \textsc{Naive-Evict} with prefix caching enabled, \textsc{vToken} reduces retained blocks by 28.6\%--42.2\%; because both variants use prefix caching, the gains come from private-suffix reclamation rather than different hit rates. Higher sharing degrees amplify this effect because protected shared-prefix blocks leave private suffixes as the main reclaimable capacity under pressure. The degree-1 case has no shared-prefix reuse and shows only the small overhead of the conservative path. A less conservative extension could use copy-on-write for shared blocks when the expected reclamation benefit justifies the extra copy.

	\subsection{Relocation Correctness and Generation Stability}
	\label{subsec:logicalkv_consistency}
	
	This experiment is the empirical test of invariants \textbf{I1} (token conservation) and \textbf{I2} (unique mapping) from \S\ref{subsec:consistency}. For each relocation event, we hash the ordered set of retained token IDs together with their K/V tensor contents and compare the hash before and after relocation; we additionally check that no logical token is lost or duplicated and that every (block, offset) referenced by the token table is valid. Across all evaluated workloads and policies, these checks hold for every relocation event, confirming that \textsc{vToken}'s indirection and asynchronous copies preserve the retained logical KV state of every request. The check is deliberately scoped to the runtime mechanism: because \textsc{Naive-Evict} and \textsc{vToken} apply identical eviction decisions, any task-quality change attributable to dropping tokens is a property of the eviction policy, not of \textsc{vToken}.
	
	Beyond exact KV-state preservation, we also check whether \textsc{vToken}'s token-table indirection and physical reclamation introduce observable generation-quality degradation. We run Llama-3.1-8B on ShareGPT under deterministic decoding, and compare \textsc{vToken} against \textsc{Naive-Evict} using 144 matched generations. As shown in Table~\ref{tab:e6-quality}, the mean paired ROUGE-L F1 difference is only $-0.0016$, and 93.1\% of the pairs differ by at most 0.01. The median \textsc{vToken}/\textsc{Naive-Evict} output-length ratio is 0.99, indicating no systematic length inflation. This paired comparison suggests that \textsc{vToken}'s indirection does not add measurable quality drift beyond the eviction policy itself.
	
	\begin{table}[t]
		\centering
		\caption{Generation stability under physical reclamation.}
		\label{tab:e6-quality}
		\footnotesize
		\setlength{\tabcolsep}{5pt}
		\renewcommand{\arraystretch}{0.98}
		\begin{tabular*}{\columnwidth}{@{}ll@{\extracolsep{\fill}}l@{}}
			\toprule
			Metric & Result & Stability \\
			\midrule
			ROUGE-L F1
			& mean $\Delta=-0.0016$
			& 93.1\% $\leq 0.01$ \\
			Output length
			& median $\textsc{vToken}/\textsc{Naive-Evict}=0.99$
			& no inflation \\
			\bottomrule
		\end{tabular*}
		\vspace{-1.6em}
	\end{table}
	
	\section{Discussion}
	\label{sec:discussion}
	
	\textsc{vToken} is a runtime-boundary abstraction rather than a vLLM-specific optimization. It targets PagedAttention-style runtimes where KV memory is managed in blocks and attention kernels access KV entries through mutable slot mappings. Porting \textsc{vToken} requires hooks for block allocation and release, pre-attention slot-map updates, and asynchronous KV copy with dependency tracking; our implementation instantiates these hooks in vLLM.
	
	\textbf{Deployment scope.} In production, \textsc{Native vLLM} remains the preferred low-pressure path, while \textsc{vToken} is a pressure-activated extension for regimes where token eviction must translate into reusable physical capacity. The capacity-frontier result measures net usable capacity under admission-controlled headroom, where bounded relocation workspace is reserved within the same KV block budget.
	
	\textbf{Scope and extensibility.}
	Our current implementation uses one runtime KV cache group and conservatively skips shared-prefix blocks to preserve prefix-cache correctness. These choices define the current implementation scope; the abstraction boundary remains the same. For MHA~\cite{vaswani2023transformer}, GQA~\cite{ainslie2023gqa}, and MQA~\cite{shazeer2019mqa} models, \textsc{vToken}'s logical token view is unchanged: the number of KV heads changes the physical tensor shape, but not token identity or slot-remapping semantics. Runtimes with multiple KV cache groups can share logical liveness state while keeping group-specific placement arrays.
	Shared-prefix support keeps private suffix blocks eligible by default and could use copy-on-write for shared blocks only when the expected reclamation benefit justifies the extra copy. In tensor-parallel settings, each TP shard has an independent KV block pool, so token-table indirection, slot remapping, and reclamation would operate on shard-local block IDs; cross-shard coordination can remain in the existing scheduler. Cross-device strategies that explicitly account for transfer cost remain orthogonal to the granularity mismatch we address.
	
	\textbf{Backend constraints and overhead.}
	Policy inputs depend on backend-exposed signals such as attention scores; the \textsc{vToken} substrate itself is independent of attention implementation. Our overhead results show that planner-side opportunity checks dominate measured overhead, while the overlap study shows that asynchronous copies avoid explicit synchronization stalls but can still contend with decode. Batching and indexing these checks within the runtime is an engineering optimization target.
	
	\section{Related Work}
	\label{sec:related}
	
	We review prior work in three areas that define the design space around \textsc{vToken}: block-based LLM inference systems, KV cache optimization mechanisms, and memory-virtualization abstractions.
	
	\textbf{LLM Inference Systems.} vLLM~\cite{kwon2023efficient} introduced PagedAttention, a block-based KV cache manager that reduces external fragmentation and enables prefix sharing. This design has been widely adopted in systems such as \textsc{TensorRT-LLM} \cite{tensorrt-llm} and \textsc{LightLLM} \cite{lightllm}. These systems provide an efficient physical substrate, but their memory-management interface remains block-oriented: allocation, reclamation, and slot mapping are expressed in terms of blocks rather than individual tokens. Other systems improve serving through parallelism, scheduling, or offloading~\cite{aminabadi2022deepspeed,sheng2023flexgen}, but do not provide a runtime abstraction that lets token-level eviction policies reclaim partially used KV blocks.
	
	\textbf{KV Cache Optimization.} KV cache optimization techniques reduce memory pressure by changing cache contents, cache representation, or the attention computation that consumes the cache. Quantization and compression approaches~\cite{hooper2024kvquant,liu2024kivi,he2024zipcache,ge2024fastgen,liu2024minicache,cai2024pyramidkv} reduce KV representation size and are orthogonal to \textsc{vToken}. Token eviction policies such as \textsc{H2O} \cite{zhang2023h2o}, \textsc{StreamingLLM} \cite{xiao2023streamingllm}, \textsc{Scissorhands} \cite{liu2023scissorhands}, and \textsc{FastGen} \cite{ge2024fastgen} decide \emph{which} tokens should be retained; their main contribution is policy design, not the memory substrate needed to realize those decisions on a block-based serving runtime.
	
	Several recent systems move closer to runtime KV management. \textsc{CacheGen} \cite{liu2024cachegen} targets compact cache transfer and \textsc{Quest} \cite{tang2024quest} reduces attention cost by selecting useful KV entries. \textsc{DiffKV}~\cite{diffkv} differentiates K/V precision, token retention, and per-head layout to compress the cache itself, requiring a mixed-precision page manager; \textsc{vToken}'s substrate operates on uniform fp16 KV and is orthogonal to such representation changes. \textsc{PagedEviction}~\cite{pagedeviction} shifts the alignment burden to the policy by making eviction page-aligned, which constrains policy-side decisions and does not directly accommodate existing token-level policies such as H2O/StreamingLLM whose importance ordering is unrelated to page boundaries; \textsc{vToken} keeps policies page-agnostic and resolves alignment in the runtime. A recent concurrent system, \textsc{Zipage}~\cite{liao2026zipage}, enforces a fixed per-request KV-block budget by relocating retained entries into bounded blocks. It is a concrete bounded-cache pipeline for reasoning workloads, whereas \textsc{vToken} exposes a policy-neutral virtualization boundary that keeps eviction policies page-agnostic and delegates slot remapping and physical reclamation to the runtime. Overall, these systems either change the cache representation, constrain eviction granularity, or build a bounded-cache pipeline; \textsc{vToken} instead preserves the block substrate and inserts the missing token-level virtualization boundary above it.
	
	\textbf{Memory Virtualization.} In LLM serving, \textsc{vAttention} \cite{vattention} virtualizes the \emph{address space} at page granularity using CUDA VMM APIs, which avoids specialized paged-attention kernels but cannot reclaim intra-page holes left by token-level eviction. \textsc{vToken} virtualizes \emph{token liveness} at token granularity above the block substrate, exposing per-token retention to the runtime so that intra-block holes can be physically reclaimed. The two layers are stackable: \textsc{vAttention} can serve as the address-space backend while \textsc{vToken} handles token-level reclamation on top. \textsc{vToken} specifically targets the missing layer between token-level eviction policies and block-based KV cache managers.
	
	\vspace{-6pt}
	\section{Conclusion}
	We introduced \textsc{vToken}, a token-level virtualization layer that makes token-level KV eviction physically effective in block-managed LLM serving runtimes. By decoupling logical token liveness from physical placement, \textsc{vToken} lets policies express which tokens to remove while the runtime handles slot remapping and physical reclamation. Our vLLM-based instantiation of \textsc{vToken} reduces retained KV blocks by 27.2\%--72.3\% and improves SLA-constrained throughput by up to 1.37\(\times\) over \textsc{Naive-Evict}; under constrained active-KV capacity, it extends the maximum feasible concurrency by up to 2$\times$. These results show that a stable logical KV view can make token-level eviction practical in existing block-based serving systems.



	\bibliographystyle{ACM-Reference-Format}
	\bibliography{sample-base}

\end{document}